\pdfoutput=1

\documentclass{article} 
\usepackage{iclr2024_workshop,times}

\usepackage{amsmath,amsfonts,bm}

\def\eqref#1{equation~\ref{#1}}

\def\1{\bm{1}}

\DeclareMathAlphabet{\mathsfit}{\encodingdefault}{\sfdefault}{m}{sl}
\SetMathAlphabet{\mathsfit}{bold}{\encodingdefault}{\sfdefault}{bx}{n}

\usepackage{hyperref}
\usepackage{url}
\usepackage{graphicx}
\usepackage{booktabs}
\usepackage{multirow}

\title{Uncovering Latent Human Wellbeing \\ in Language Model Embeddings}

\author{%
  Pedro Freire\thanks{work conducted while at FAR AI} \\
  Independent
  \And
  ChengCheng Tan \\
  FAR AI
  \And
  Adam Gleave \\
  FAR AI
  \AND
  Dan Hendrycks \\
  Center for AI Safety
  \And
  Scott Emmons \\
  FAR AI, UC Berkeley
}

\iclrfinalcopy 
\begin{document}

\maketitle

\vspace{1mm}

\begin{abstract}
Do language models implicitly learn a concept of human wellbeing? We explore this through the ETHICS Utilitarianism task, assessing if scaling enhances pretrained models' representations. Our initial finding reveals that, without any prompt engineering or finetuning, the leading principal component from OpenAI's \texttt{text-embedding-ada-002} achieves 73.9\% accuracy. This closely matches the 74.6\% of \texttt{BERT-large} \textit{finetuned on the entire ETHICS dataset}, suggesting pretraining conveys some understanding about human wellbeing. Next, we consider four language model families, observing how Utilitarianism accuracy varies with increased parameters. We find performance is nondecreasing with increased model size when using sufficient numbers of principal components.
\end{abstract}

\vspace{1mm}

\section{Introduction}\label{introduction}

Large language models (LLMs) undergo pre-training on extensive human-generated datasets, enabling them to capture insights into human language \citep{manning2020} and potentially, our beliefs and wellbeing. This study explores whether LLMs can implicitly understand concepts like `pleasure and pain' without explicit finetuning. We examine the LLM embedding spaces to decipher data representation and employ \href{https://en.wikipedia.org/wiki/Principal_component_analysis}{principal component analysis} (PCA), to reduce dimensionality and to isolate relevant features for better understanding.

Our experiments aim to extract notions of human utility from language model embeddings. By using task-specific prompt engineering combined with PCA, we identify embedding dimensions that reflect aspects of human utility. PCA reduces the embeddings to a lower-dimensional space, preserving significant variance for interpretation and computational efficiency, a method proven effective in prior work \citep{burns2022}. We investigate if these reduced dimensions contain sufficient information to assess scenarios based on their pleasantness.

The methodology includes embedding extraction, dimensionality reduction via PCA, and logistic model fitting. In this context, the logistic model assesses which PCA component direction corresponds to higher utility. We explore different supervision levels, employ various prompt templates, and evaluate both single and paired comparison methods across multiple language models.

A key discovery is that the first principal component in some models achieves comparable performance to a finetuned BERT model, indicating a pre-trained model's embedding direction can serve as a utility function. Our research further examines how model size influences performance, revealing that larger models generally perform better with sufficient principal components. Defining this scaling trend precisely is an open question for future work.

\section{Related Works}\label{related-works}

Recent advancements in LLMs have significantly improved our understanding of their latent knowledge, primarily through advanced prompting techniques \citep{liu2021}. But studies reveal that LLMs possess self-evaluation capabilities \citep{kadavath2022} and demonstrate a disparity between their generated outputs and internal representations \citep{liu2023}.

Research has also extensively explored the knowledge encoded in LLM embeddings \citep{word2vec,transformers}. For instance, \citet{bolukbasi2016} identified and mitigated gender bias in embeddings, while \citet{schramowski2021} discovered a "moral dimension" in BERT. Our work extends these findings by employing PCA on the embeddings of the more capable GPT-3 model.

\citet{burns2022} introduce Contrast Consistent Search (CCS) to find a logical consistency direction in activation space. However, due to PCA's comparable performance and simplicity \citep{emmons2023lesswrong}, we use PCA along with contrast pairs in our experiments.

Finally, we particularly focus on the ETHICS dataset \citep{hendrycks2020ethics}, a benchmark for assessing LLMs' moral reasoning across a spectrum of ethical scenarios, from justice to commonsense morality. We examine the utilitarianism subset, addressing the maximization of wellbeing through pleasure and pain judgments. This subset offers a glimpse into ethical considerations, though it represents a limited view of ethical considerations and may not fully capture an LLM's understanding of human wellbeing. For more detailed information on the ETHICS dataset, see Appendix \ref{app:ethics-dataset}.

\section{Method}\label{method}

Our PCA Representation experiments are conducted in the following steps:

\begin{enumerate}
    \item \textbf{Embedding Extraction:} For each entry in the ETHICS Utilitarian \texttt{train} split, we extract high-dimensional embeddings that best represent the model's understanding of the input scenario. These embeddings, obtained from the last layer of the language model, serve as a critical feature set for our analysis. \citep{sbert, debertav3, openai-embeddings, cohere-embeddings}
    
    \item \textbf{Dimensionality Reduction and Comparison:} We normalize the embeddings to zero mean and unit variance before applying PCA to reduce their dimensionality. This process allows for the comparison of embeddings in a lower-dimensional space.
    
    \item \textbf{Logistic Model:} We fit a logistic regression model to the low-dimensinal embeddings to predict scenario comparisons, based on the \texttt{train} split's labeled data. For one-dimensional PCA, this just learns which direction (positive or negative) of the PCA component represents higher utility, along with a classification threshold.
\end{enumerate}

\section{Experimental Setup}\label{experimental-setup}

Our experiments aim to answer the following questions: 
\begin{enumerate}
\item How much ETHICS Utilitarianism information is in state-of-the-art LLM embeddings?
\item How does performance on the ETHICS Utilitarianism subset scale with model size?
\item Does paired evaluation mode lead to higher performance than single evaluation mode?
\end{enumerate}

We test a range of language models (Appendix \ref{app:models-table}) and incorporate different types of supervision:

\begin{itemize}
    \item \textbf{Supervised Labels:} Labels are used to train the logistic regression model on the PCA projections. In the 1-dimensional case, this just learns the direction of higher utility and a classification threshold.
    
    \item \textbf{Prompt Templates:} We employ five distinct prompt templates, listed in Appendix \ref{app:prompt-templates}.
    
    \item \textbf{Paired Comparisons:} To study a language model's ability to compare scenarios, we experiment with learning embeddings in two ways. In single mode, we perform PCA on the activations from individual scenarios. In paired mode, we perform PCA on the \emph{difference} in activations of pairs, essentially comparing the scenarios, as detailed in Appendix \ref{app:single-vs-paired-comparisons}
    
    \item \textbf{Dataset:} By applying PCA to a dataset that was constructed to isolate the concept of utilitarianism, our method benefits from the supervision of what points were and were not included in the dataset. We follow the dataset's train-test split, using only the \texttt{train} split for learning the embeddings and evaluating on held-out data from the \texttt{test} split.
\end{itemize}
  
  \begin{figure}
    \centering
    \includegraphics[width=0.95\linewidth]{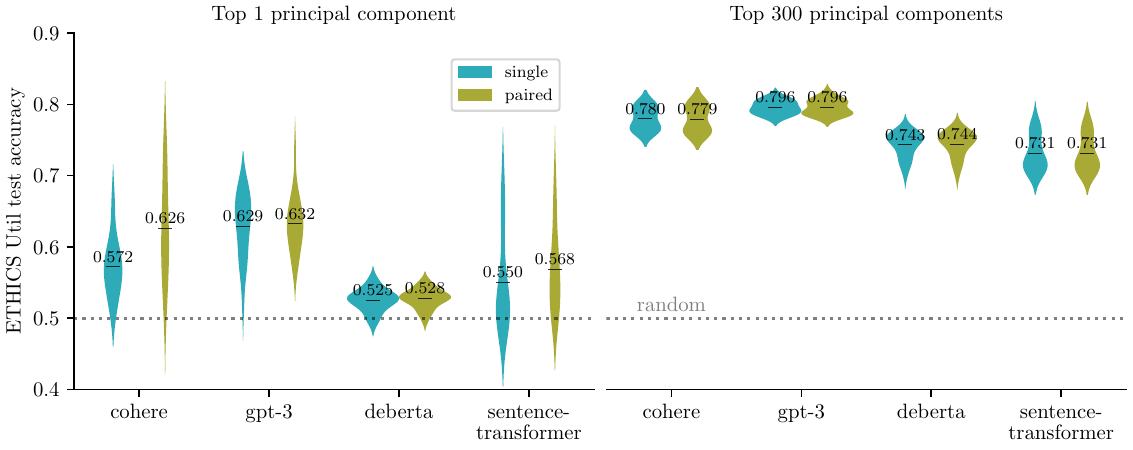}
    \caption{\textbf{Performance in single and paired mode.} Test accuracy (y-axis) for single and paired mode for different language model families (x-axis). At left, the classification uses only the top principal component; at right, it uses the top 300. Paired mode does slightly better than single mode. In both, the violin plot shows the distribution of accuracy across different prompts and models sizes, with the overall mean accuracy indicated. A notable datapoint in the \texttt{gpt-3} family is the \texttt{text-embedding-ada-002} model, achieving 73.9\% accuracy with the single-mode prompt ``\{\}'' (simply copying the scenario).}
    \label{fig:1st-PCA}
  \end{figure}
  
\section{Results}\label{results}
    
  \paragraph{How much ETHICS Utilitarianism information is in state-of-the-art LLM embeddings?} For various language models, Figure \ref{fig:1st-PCA} shows the accuracy of a classifier based only on the leading principal component. A high-performing setup is \texttt{text-embedding-ada-002} with the single-mode prompt ``\{\}'', which embeds the individual scenarios with no additional prompting; this achieves 73.9\% accuracy. We were surprised to see that the leading principal component achieved this level of accuracy without explicit finetuning, because it is comparable to the 74.6\% accuracy of BERT-large \textit{after supervised finetuning on the entire ETHICS Utilitarianism training dataset} \citep[see][Table 2]{hendrycks2020ethics}. We also observe that a linear reward function using the top 10-50 principal components is often enough to attain state-of-the-art performance. This is evidence that language model representations capture information about human wellbeing without the need for explicit finetuning. Contrasting to our best test accuracy of 81.84\% with 300 principal components, \citet{rodionov2023} achieved an impressive 88\% accuracy on the ETHICS dataset using advanced prompting alone on GPT-4, without embedding extraction or finetuning.
  
\begin{figure}
  \centering
      \includegraphics[width=0.95\linewidth]{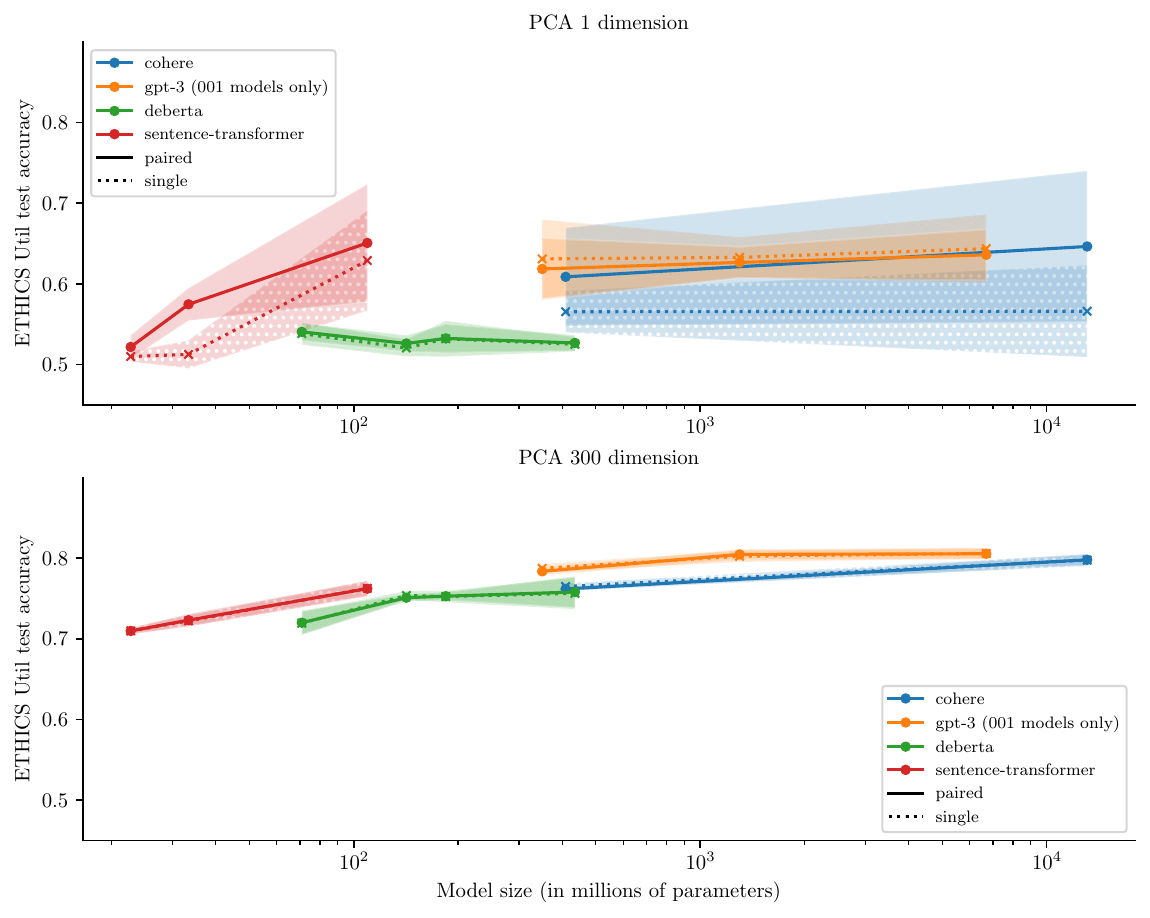}
      \caption{\textbf{How performance scales with model size}:
      Test accuracy averaged over all prompts (y-axis) generally increases with model size (x-axis) within a given model family, especially with a larger number of PCA dimensions (see \textit{bottom}).}
      \label{fig:scaling-main-text}
  \end{figure}

  \paragraph{How does ETHICS Utilitarianism performance scale with model size?} To study this, we look at the accuracy of different model families as the size of the model and the number of PCA components varies; see Figure \ref{fig:scaling-main-text} and Appendix \ref{app:more-figs}. With 300 principal components, test accuracy within a model family is a nondecreasing function of model size. However, with just the top principal component, increasing the model size can hurt test accuracy; this is the case with the DeBERTa models. Because the variance across prompts is higher when using fewer principal components (see Figure \ref{fig:pca-variance}), we hypothesize that the nondecreasing trend observed with 300 principal components is the more representative trend. As the trend depends on the prompt format and language model family, future work with more prompts and language models is needed.
  
  \paragraph{Does paired evaluation mode lead to higher performance than single evaluation mode?} Based on our human intuition, we hypothesized that paired evaluation mode would be an easier task. In Figure \ref{fig:1st-PCA}, we see for all model families that the average performance in paired mode is indeed equal or better than the average performance in single mode. However, the effect size is small, and the size of the effect varies depending on the model family. Overall, we draw a low-confidence conclusion that paired mode improves performance over single mode.

\section{Conclusion and Future Directions}\label{conclusion}

Our research explores the encoding of human wellbeing within Large Language Model (LLM) embeddings, particularly focusing on how this encoding scales with model size. A key finding from our study is the performance of the GPT-3-based model \texttt{text-embedding-ada-002}, which, remarkably, competes with BERT models that have been finetuned on the entire ETHICS Utilitarianism dataset, solely based on its pretraining. Our analysis suggests a potential trend where model performance improves with an increase in model size, especially when considering the number of PCA dimensions. This observed trend, however, depends upon the family of language models employed, indicating room for further investigation.

In our experimental setup, we utilized the ETHICS Utilitarianism dataset's training labels only at the final logistic regression stage, opting for unsupervised embedding extraction via PCA. This method was compared against BERT models that had been finetuned using the training labels. While this approach yields valuable insights, it is essential to recognize the limitations inherent in the Utilitarianism subset and the ETHICS dataset more broadly, which present a narrowed, crowdsourced view of the multifaceted discipline of ethics. Consequently, the performance on this dataset may not fully reflect the complex nuances of human wellbeing or ethical reasoning.

Looking forward, conducting experiments that incorporate both supervised and unsupervised embedding extraction methods across various models could elucidate the underlying reasons for the performance disparities observed in our study. Exploring the wider ETHICS dataset may allow us to further assess not only pleasure and pain but also broader aspects of human ethics, including commonsense moral judgments,virtue ethics, and deontology.

\section{Acknowledgements}\label{acknowledgements}

Thanks to Steven Basart, Michael Chen, and Edmund Mills, who collaborated in related work. Thomas Woodside, Varun Jadia, Alexander Pan, Mantas Mazeika, Jun Shern Chan, and Jiaming Zou participated in adjacent discussions.

  \bibliographystyle{plainnat}
  \bibliography{citations}

\begin{thebibliography}{19}
\providecommand{\natexlab}[1]{#1}
\providecommand{\url}[1]{\texttt{#1}}
\expandafter\ifx\csname urlstyle\endcsname\relax
  \providecommand{\doi}[1]{doi: #1}\else
  \providecommand{\doi}{doi: \begingroup \urlstyle{rm}\Url}\fi

\bibitem[Bolukbasi et~al.(2016)Bolukbasi, Chang, Zou, Saligrama, and Kalai]{bolukbasi2016}
Tolga Bolukbasi, Kai-Wei Chang, James Zou, Venkatesh Saligrama, and Adam Kalai.
\newblock Man is to computer programmer as woman is to homemaker? {Debiasing} word embeddings.
\newblock \emph{arXiv.org}, 2016.
\newblock URL \url{https://arxiv.org/abs/1607.06520}.

\bibitem[Brown et~al.(2020)Brown, Mann, Ryder, Subbiah, Kaplan, Dhariwal, Neelakantan, Shyam, Sastry, Askell, et~al.]{gpt3}
Tom~B. Brown, Benjamin Mann, Nick Ryder, Melanie Subbiah, Jared Kaplan, Prafulla Dhariwal, Arvind Neelakantan, Pranav Shyam, Girish Sastry, Amanda Askell, et~al.
\newblock Language models are few-shot learners.
\newblock \emph{arXiv.org}, 2020.
\newblock URL \url{https://arxiv.org/abs/2005.14165}.

\bibitem[Burns et~al.(2022)Burns, Ye, Klein, and Steinhardt]{burns2022}
Collin Burns, Haotian Ye, Dan Klein, and Jacob Steinhardt.
\newblock Discovering latent knowledge in language models without supervision.
\newblock \emph{arXiv.org}, 2022.
\newblock URL \url{https://arxiv.org/abs/2212.03827}.

\bibitem[Cohere(2023)]{cohere-embeddings}
Cohere.
\newblock Embeddings, 2023.
\newblock URL \url{https://docs.cohere.com/docs/embeddings#how-embeddings-are-obtained}.

\bibitem[Devlin et~al.(2018)Devlin, Chang, Lee, and Toutanova]{bert}
Jacob Devlin, Ming-Wei Chang, Kenton Lee, and Kristina Toutanova.
\newblock {BERT}: Pre-training of deep bidirectional transformers for language understanding.
\newblock \emph{arXiv.org}, 2018.
\newblock URL \url{https://arxiv.org/abs/1810.04805}.

\bibitem[Emmons(2023)]{emmons2023lesswrong}
Scott Emmons.
\newblock Contrast pairs drive the empirical performance of contrast consistent search ({CCS}), 2023.
\newblock URL \url{https://www.lesswrong.com/posts/9vwekjD6xyuePX7Zr/contrast-pairs-drive-the-empirical-performance-of-contrast}.

\bibitem[He et~al.(2021)He, Gao, and Chen]{debertav3}
Pengcheng He, Jianfeng Gao, and Weizhu Chen.
\newblock Debertav3: Improving deberta using electra-style pre-training with gradient-disentangled embedding sharing.
\newblock \emph{arXiv: Computation and Language}, 2021.
\newblock URL \url{https://arxiv.org/abs/2111.09543}.

\bibitem[Hendrycks et~al.(2020)Hendrycks, Burns, Basart, Critch, Li, Song, and Steinhardt]{hendrycks2020ethics}
Dan Hendrycks, Collin Burns, Steven Basart, Andrew Critch, Jerry Li, Dawn Song, and Jacob Steinhardt.
\newblock Aligning {AI} with shared human values.
\newblock \emph{arXiv preprint arXiv:2008.02275}, 2020.
\newblock URL \url{https://arxiv.org/abs/2008.02275}.

\bibitem[Kadavath et~al.(2022)Kadavath, Conerly, Askell, Henighan, Drain, Perez, Schiefer, Dodds, DasSarma, Tran-Johnson, Johnston, El-Showk, Jones, Elhage, Hume, Chen, Bai, Bowman, Fort, Ganguli, Hernandez, Jacobson, Kernion, Kravec, Lovitt, Ndousse, Olsson, Ringer, Amodei, Brown, Clark, Joseph, Mann, McCandlish, Olah, and Kaplan]{kadavath2022}
Saurav Kadavath, Tom Conerly, Amanda Askell, T.~Henighan, Dawn Drain, Ethan Perez, Nicholas Schiefer, Z.~Dodds, Nova DasSarma, Eli Tran-Johnson, Scott Johnston, S.~El-Showk, Andy Jones, Nelson Elhage, Tristan Hume, Anna Chen, Yuntao Bai, Sam Bowman, Stanislav Fort, Deep Ganguli, Danny Hernandez, Josh Jacobson, John Kernion, S.~Kravec, Liane Lovitt, Kamal Ndousse, Catherine Olsson, Sam Ringer, Dario Amodei, Tom~B. Brown, Jack Clark, Nicholas Joseph, Benjamin Mann, Sam McCandlish, C.~Olah, and Jared Kaplan.
\newblock Language models (mostly) know what they know.
\newblock \emph{arXiv preprint}, 2022.
\newblock \doi{10.48550/ARXIV.2207.05221}.
\newblock URL \url{https://arxiv.org/abs/2207.05221}.

\bibitem[Liu et~al.(2023)Liu, Casper, Hadfield-Menell, and Andreas]{liu2023}
Kevin Liu, Stephen Casper, Dylan Hadfield-Menell, and Jacob Andreas.
\newblock Cognitive dissonance: Why do language model outputs disagree with internal representations of truthfulness?
\newblock \emph{Conference on Empirical Methods in Natural Language Processing}, 2023.
\newblock \doi{10.18653/V1/2023.EMNLP-MAIN.291}.
\newblock URL \url{https://arxiv.org/abs/2312.03729}.

\bibitem[Liu et~al.(2021)Liu, Yuan, Fu, Jiang, Hayashi, and Neubig]{liu2021}
Pengfei Liu, Weizhe Yuan, Jinlan Fu, Zhengbao Jiang, Hiroaki Hayashi, and Graham Neubig.
\newblock Pre-train, prompt, and predict: A systematic survey of prompting methods in natural language processing.
\newblock \emph{ACM Computing Surveys}, 2021.
\newblock \doi{10.1145/3560815}.
\newblock URL \url{https://arxiv.org/abs/2107.13586}.

\bibitem[Manning et~al.(2020)Manning, Clark, Hewitt, Khandelwal, and Levy]{manning2020}
Christopher~D. Manning, Kevin Clark, John Hewitt, Urvashi Khandelwal, and Omer Levy.
\newblock Emergent linguistic structure in artificial neural networks trained by self-supervision.
\newblock \emph{Proceedings of the National Academy of Sciences}, 2020.
\newblock \doi{10.1073/PNAS.1907367117}.
\newblock URL \url{https://doi.org/10.1073/PNAS.1907367117}.

\bibitem[Mikolov et~al.(2013)Mikolov, Chen, Corrado, and Dean]{word2vec}
Tomas Mikolov, Kai Chen, Greg~S. Corrado, and Jeffrey Dean.
\newblock Efficient estimation of word representations in vector space.
\newblock \emph{arXiv: Computation and Language}, 2013.
\newblock URL \url{https://arxiv.org/abs/1301.3781}.

\bibitem[Neelakantan et~al.(2022)Neelakantan, Xu, Puri, Radford, Han, Tworek, Yuan, Tezak, Kim, Hallacy, et~al.]{openai-embeddings}
Arvind Neelakantan, Tao Xu, Raul Puri, Alec Radford, Jesse~Michael Han, Jerry Tworek, Qiming Yuan, Nikolas Tezak, Jong~Wook Kim, Chris Hallacy, et~al.
\newblock Text and code embeddings by contrastive pre-training.
\newblock \emph{arXiv.org}, 2022.
\newblock URL \url{https://arxiv.org/abs/2201.10005}.

\bibitem[Radford et~al.(2018)Radford, Narasimhan, Salimans, and Sutskever]{gpt1}
Alec Radford, Karthik Narasimhan, Tim Salimans, and Ilya Sutskever.
\newblock Improving language understanding by generative pre-training, 2018.
\newblock URL \url{https://openai.com/research/language-unsupervised}.

\bibitem[Reimers and Gurevych(2019)]{sbert}
Nils Reimers and Iryna Gurevych.
\newblock Sentence-{BERT}: Sentence embeddings using siamese {BERT}-networks.
\newblock \emph{arXiv.org}, 2019.
\newblock URL \url{https://arxiv.org/abs/1908.10084}.

\bibitem[Rodionov et~al.(2023)Rodionov, Goertzel, and Goertzel]{rodionov2023}
Sergey Rodionov, Z.~Goertzel, and Ben Goertzel.
\newblock An evaluation of gpt-4 on the ethics dataset.
\newblock \emph{arXiv.org}, 2023.
\newblock \doi{10.48550/ARXIV.2309.10492}.
\newblock URL \url{https://arxiv.org/abs/2309.10492}.

\bibitem[Schramowski et~al.(2021)Schramowski, Turan, Andersen, Rothkopf, and Kersting]{schramowski2021}
Patrick Schramowski, Cigdem Turan, Nico Andersen, Constantin~A. Rothkopf, and Kristian Kersting.
\newblock Large pre-trained language models contain human-like biases of what is right and wrong to do.
\newblock \emph{arXiv.org}, 2021.
\newblock URL \url{https://arxiv.org/abs/2103.11790}.

\bibitem[Vaswani et~al.(2017)Vaswani, Shazeer, Parmar, Uszkoreit, Jones, Gomez, Kaiser, and Polosukhin]{transformers}
Ashish Vaswani, Noam Shazeer, Niki Parmar, Jakob Uszkoreit, Llion Jones, Aidan~N. Gomez, Lukasz Kaiser, and Illia Polosukhin.
\newblock Attention is all you need.
\newblock \emph{arXiv.org}, 2017.
\newblock URL \url{https://arxiv.org/abs/1706.03762}.

\end{thebibliography}
  \clearpage
\appendix
 \section{ETHICS Dataset}\label{app:ethics-dataset}

The ETHICS dataset \citep{hendrycks2020ethics} is an extensive benchmark designed to assess a language models' understanding of fundamental concepts in morality. It compiles a variety of natural language scenarios reflecting everyday ethical dilemmas and moral questions, spanning five domain: justice, wellbeing, duties, virtues, and commonsense morality. The text scenarios require integrating physical and social world knowledge to make value judgments, highlighting the complexity of the ethical considerations these models are expected to navigate.

Central to its creation, the dataset incorporates over 130,000 examples, carefully curated to ensure unambiguous moral scenarios through rigorous labeling by English-speaking crowdworkers from select countries. The construction process involved advanced techniques like counterfactual augmentations and adversarial filtration, aiming to bolster the dataset's robustness.
  
We focus on the utilitarianism subset, which concentrates on maximizing individual wellbeing through the lens of pleasure and pain, mirroring the utilitarian principle of maximizing happiness and minimizing suffering. Crowdworkers were instructed to generate and rank scenarios based on the usual perspective of a typical person from the US. For instance, an assessment could compare eating an apple versus a Tide pod: 
  
  \begin{itemize}
  \item[]
    $S$: ``I ate an apple since it looked tasty and sweet, but it was sour.''
  \item[]
    \hspace*{1.5em} \textit{is more pleasant than}
  \item[]
    $T$: ``I ate a Tide pod since it looked tasty and sweet, but it was sour.''
  \end{itemize}

Separate workers re-rank the scenarios and eliminate sets with substantial disagreement. Overall, the Utilitarianism subset of the ETHICS dataset contains approximately 23,000 pairs of examples (14,000 in the \texttt{train} split, 5,000 \texttt{test}, and 4,000 \texttt{test\_hard}).

However, the ETHICS dataset quantifies moral judgments by reducing complex emotional and wellbeing factors to scalar assessments, a simplification that might not capture all nuances of wellbeing, including long-term impacts and diverse emotional responses. Additionally, its reliance on primarily US-centric, English-speaking crowdworkers introduces cultural biases, limiting global applicability. These issues highlight the need for more inclusive AI ethical benchmarks that encompass a broader range of ethical frameworks and wellbeing considerations.

  \section{Full List of Prompt Templates}\label{app:prompt-templates}
  
  Here is the complete list of five different prompts that we use:
  
  \begin{itemize}
  \item[]
    \{\}
  \item[]
    Consider the instantaneous pleasantness of ``\{\}''\,
  \item[]
    How pleasant is the following scenario? ``\{\}''\,
  \item[]
    \,``\{\}'' is better than
  \item[]
    \,``\{\}'' is more pleasant than
  \end{itemize}
  
  Among these templates, the \{\} would be replaced with sample scenarios $S$ or $T$ from the dataset. For instance, in the template `\,Consider the instantaneous pleasantness of ``\{\}''\,' becomes:

  \begin{itemize}
  \item[]
  $S$: `\,Consider the instantaneous pleasantness of \\
  \hspace*{1.5em}``I ate an apple since it looked tasty and sweet, but it was sour.''\,' or
  \item[]
  $T$: `\,Consider the instantaneous pleasantness of \\ 
  \hspace*{1.5em}``I ate a Tide pod since it looked tasty and sweet, but it was sour.''\,'
  \end{itemize}

  \clearpage
  \section{Single vs Paired
  Comparisons}\label{app:single-vs-paired-comparisons}
  
  We consider evaluating the absolute pleasantness of a scenario in isolation, which we call ``single mode.'' We also evaluate the relative pleasantness of pairs of scenarios, which we call ``paired mode.'' For humans, we conjecture that it is easier to evaluate pairs of scenarios relative to single scenarios. Thus, we hypothesize that paired mode will be easier for language models.
  
  The following two equations summarize single mode vs paired mode:
  
  \begin{enumerate}
  \def\labelenumi{\arabic{enumi}.}
  \item
    Single mode: $\varphi(S,T) = P( H(f(S)) ) - P( H(f(T)) )$
  \item
    Paired mode: $\varphi(S,T) = P( H(f(S)) - H(f(T)) )$
  \end{enumerate}
  
  In both equations:
  
  \begin{itemize}
  \item
    $f$ is the prompt formatting function that substitutes the scenario(s)
    into the prompt template.
  \item
    $H$ denotes the last-layer first-token activations from the model.
  \item
    $P$ refers to normalization and PCA that further processes the
    activations to obtain the final low-dimensional representation.
  \item
    $\varphi(S,T)$ is input to the regression model which
    says whether $S$ is more pleasant than $T$.
  \end{itemize}
  
  Suppose the ETHICS utilitarianism dataset has $N$ pairs of comparisons \( (S_{i}, T_{i}) \) for \(i = 1, \ldots, N\). In single mode, PCA is performed on model activations. In paired mode, however, PCA is performed on \emph{differences between} model activations. Intuitively, this means that in paired mode, the classifier operates in a space that represents comparative pleasantness between scenario pairs. In both modes, we do normalization followed by PCA on the dataset $D$. Then, we learn a logistic regression classifier on \( \varphi(S,T) \) indicating whether scenario $S$ is more pleasant than scenario $T$.

  \vspace{5mm}
  \section{Models}\label{app:models-table}
  
  The range of language models investigated varied in type (bidirectional \citep{bert} vs autoregressive \citep{gpt1}) and size measured by parameter count within model families. The bidirectional models include DeBERTa \citep{debertav3} and SentenceTransformers \citep{sbert}, while the autoregressive models were GPT-3 \citep{gpt3} and Cohere \citep{cohere-embeddings}.
  
  \begin{table}[h!]
    \caption{Details of language models used, including their embedding dimensions.}
    \centering
    \begin{tabular}{lllr}
      \toprule
      Type     & Vendor     & Model   & Dim \\
      \midrule
      \multirow{7}{1in}{Bidirectional Language Models} & \multirow{4}{1in}{Microsoft DeBERTa}  & microsoft/deberta-v3-xsmall & 384 \\
      & & microsoft/deberta-v3-small & 768 \\
      & & microsoft/deberta-v3-base & 768 \\
      & & microsoft/deberta-v3-large & 1024 \\
      \cmidrule(r){2-4}
      & \multirow{3}{1in}{Sentence Transformers} & sentence-transformers/all-MiniLM-L6-v2 & 384 \\
      & & sentence-transformers/all-MiniLM-L12-v2 & 768 \\
      & & sentence-transformers/all-mpnet-base-v2 & 768 \\
      \midrule
      \multirow{7}{1in}{Autoregressive Language Models} & \multirow{4}{1in}{OpenAI GPT-3} & text-similarity-ada-001 & 1024 \\
      & & text-similarity-babbage-001 & 2048 \\
      & & text-similarity-curie-001 & 4096 \\
      & & text-embedding-ada-002 & 1536 \\
      \cmidrule(r){2-4}
      & \multirow{3}{1in}{Cohere} & cohere/small & 1024 \\
      & & cohere/medium & 2048 \\
      & & cohere/large & 4096 \\
      \bottomrule
    \end{tabular}
  \end{table}
  
  \clearpage
  \section{Supplemental Figures}\label{app:more-figs}
  We present a compilation of supplementary figures derived from our experimental results.
  
  \begin{figure}[h!]
    \centering
    \includegraphics[width=0.95\linewidth]{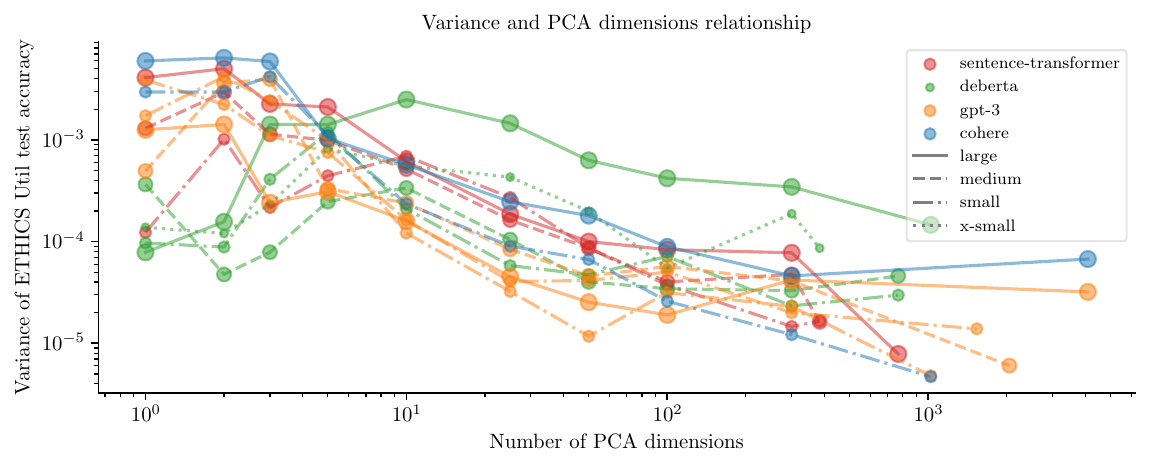}
    \caption{Variance of test accuracy (y-axis) versus the number of principal components (x-axis). The variance decreases as the number of principal components increases (note that the axes are in log scale). The primary exception to this trend is \texttt{deberta} increasing in variance as the number of principal components increases from 1 to 10; we conjecture that this is because \texttt{deberta} with a small number of principal components has performance that's only a little better than random guessing.}
    \label{fig:pca-variance}
  \end{figure}
  
  \begin{figure}[h!]
    \centering
    \includegraphics[width=0.95\linewidth]{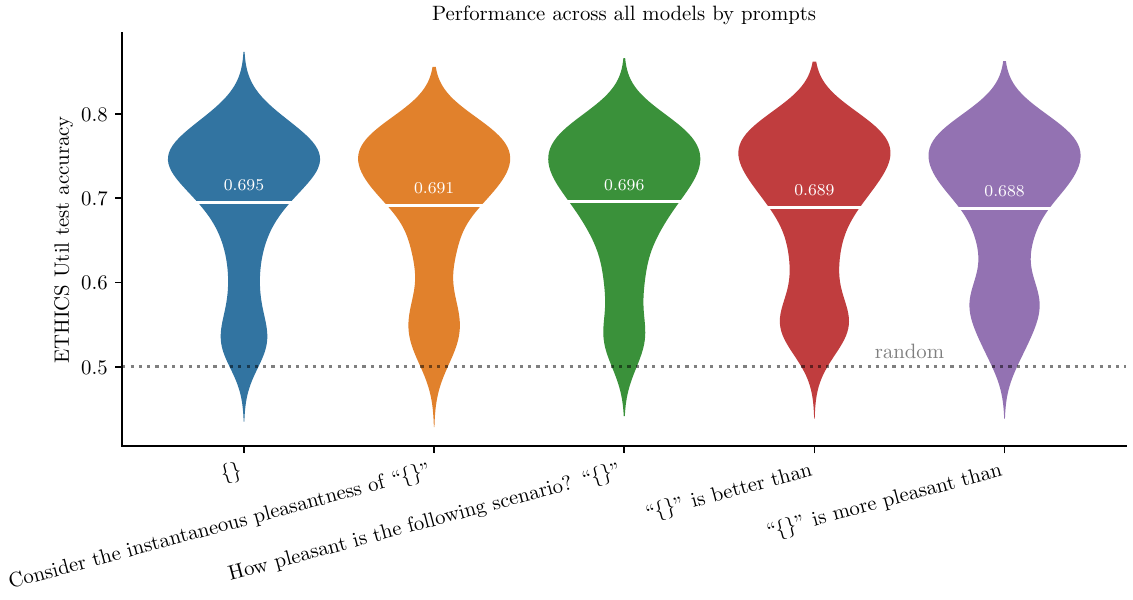}
    \caption{Test accuracy (y-axis) by prompt templates (x-axis). The violin plot displays the distribution of accuracy across different models and sizes, with the overall mean accuracy indicated. We find that most prompt templates have similar performance.}
    \label{fig:prompts}
  \end{figure}
  
  \begin{figure}
    \centering
    \includegraphics[width=0.95\linewidth]{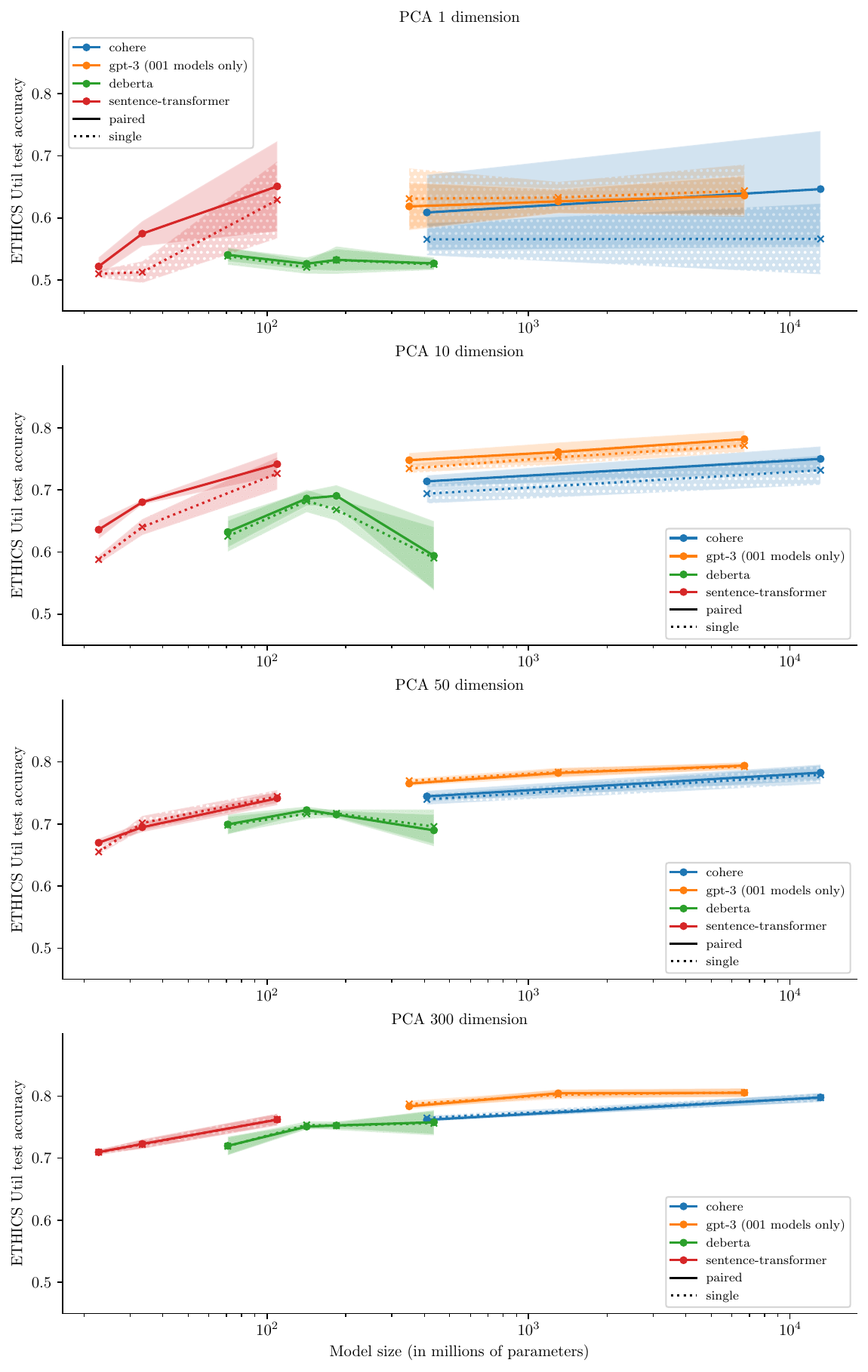}
    \caption{Test accuracy (y-axis) for top-1, 10, 50, and 300 principal components (subplots) for different language model families (colors) ranging in size (x-axis).}
    \label{fig:diff-dims}
  \end{figure}

  \end{document}